# Robust Image Classification in the Presence of Out-of-Distribution and Adversarial Samples Using Attractors in Neural Networks

**Nasrin Alipour , Seyyed Ali SeyyedSalehi** [*]

**Biomedical Engineering Department, Amirkabir University of Technology, Tehran, Iran**

**Abstract**

The proper handling of out-of-distribution (OOD) samples in deep classifiers is a critical concern for ensuring the suitability of deep neural networks in safety-critical systems. Existing approaches developed for robust OOD detection in the presence of adversarial attacks lose their performance by increasing the perturbation levels. This study proposes a method for robust classification in the presence of OOD samples and adversarial attacks with high perturbation levels. The proposed approach utilizes a fully connected neural network that is trained to use training samples as its attractors, enhancing its robustness. This network has the ability to classify inputs and identify OOD samples as well. To evaluate this method, the network is trained on the MNIST dataset, and its performance is tested on adversarial examples. The results indicate that the network maintains its performance even when classifying adversarial examples, achieving 87.13% accuracy when dealing with highly perturbed MNIST test data. Furthermore, by using fashion-MNIST and CIFAR-10-bw as OOD samples, the network can distinguish these samples from MNIST samples with an accuracy of 98.84% and 99.28%, respectively. In the presence of severe adversarial attacks, these measures decrease slightly to 98.48% and 98.88%, indicating the robustness of the proposed method.

**Key Words:**
**OOD detection, Robustness, Attractor, autoencoder networks, image classification**

# 1 Introduction

Despite the widespread application of deep neural networks across various domains, concerns persist regarding their suitability for safety-critical systems like medical diagnostics and autonomous vehicles. Studies have highlighted several unexpected failure modes of deep learning classifiers, including susceptibility to natural perturbations [1], inclination towards overconfident predictions [2,3], and vulnerability to adversarial attacks [4]. When the training and test distributions differ in real-world tasks, neural network classifiers tend to fail, especially networks that use the ReLU activation function [5]. This sometimes leads to confident yet incorrect predictions, known as silent failures [6]. Such errors can have serious consequences, particularly in critical domains.

To address this problem, an out-of-distribution (OOD) detector helps distinguish whether an input belongs to the training data distribution, known as in-distribution (ID) examples, or a different one (OOD examples). Recent studies have shown that common OOD detection methods, such as MSP [3], ODIN [7], Mahalanobis [8], and OE [22], are also vulnerable to adversarial perturbations, leading to decreased performance. This highlights the significance of developing robust and reliable OOD detection methods, as emphasized by several research works [5,9,10,11]. Specifically, robustness against adversarially selected inputs is emerging as a vital objective in developing neural networks. Although trained models succeed in classifying benign inputs, recent studies indicate that adversarial attacks can manipulate inputs that can cause the network to produce incorrect outputs, often without being perceptible to humans [2, 12, 13]. The efficiency of these attacks, as well as the robustness of neural networks against them, is a critical area of research.

---

[*] Corresponding author. Address: Hafez St., Valiasr Square, Tehran 1591634311, Iran.
E-mail addresses: nasi@aut.ac.ir (N. Alipour), ssalehi@aut.ac.ir (S.A. Seyyedsalehi).

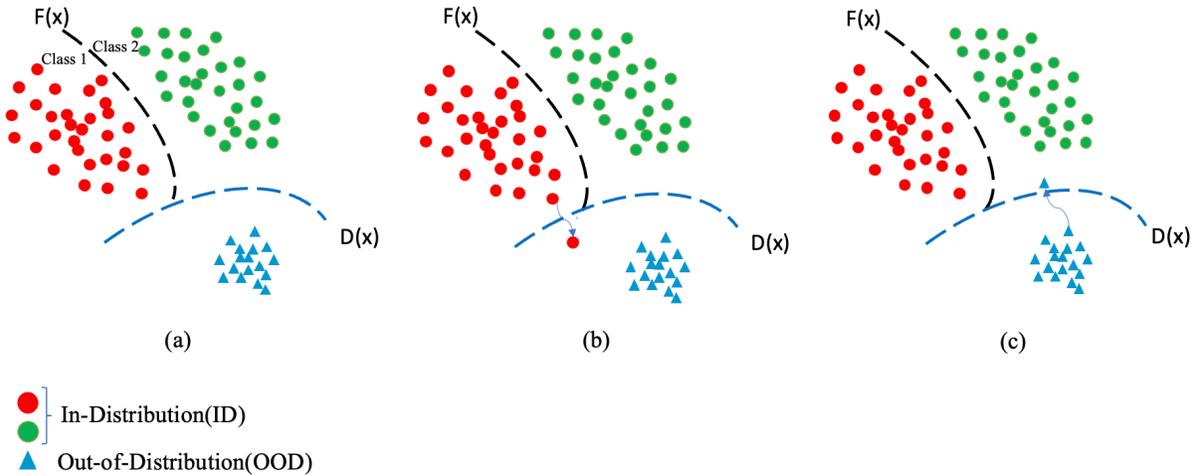

**Figure 1:** Classification in the presence of OOD and adversarial examples. The red and green points are ID examples and illustrate different classes learned by classifier F(x), and the blue triangles are from a different distribution. (a) Performing OOD detection using D(x) to prevent classifier F(x) from incorrectly classifying OOD examples, (b) The capability of adversarial ID examples to deceive D(x), (c) The capability of adversarial OOD examples to deceive D(x).

White box adversarial attacks, such as the Fast Gradient Sign Method (FGSM) [14] and Projected Gradient Descent (PGD) [15], pose significant challenges to the robustness of neural networks, and that is why the research community is dedicated to discovering effective solutions for this issue. [16, 17].

If we represent the classifier as F(x) and the OOD detector as D(x), then the OOD detector's role is to distinguish between ID and OOD examples. By rejecting OOD examples, D(x) helps prevent the F(x) from making incorrect classifications when encountering such examples. This process is illustrated in Figure 1-a. Adversarial attacks can deceive both D(x) and F(x), leading to incorrect input class prediction. In two cases, D(x) can be fooled by adversarial examples. First, when the adversary constructs adversarial ID examples by adding small perturbations to the normal ID inputs such that the D(x) will falsely reject them (Figure 1-b). Second, when the adversarial OOD examples are generated by manipulating the OOD inputs, the OOD detector will fail to reject them (Figure 1-c). Even after successful OOD detection, adversarial attacks can cause F(x) to have incorrect predictions, ultimately weakening the overall classification system.

The human brain is capable of handling a wide range of input variations. However, neural networks inspired by the functionality of neurons in the human brain lack this capability, and perturbations can affect their decisions. This shows that the cognition processes in neural networks and the human brain differ. Some evidence suggests that the human brain's connections are bidirectional [18], acting like an attractor when memorizing or retrieving information. Until today, some studies have focused on the attractors in neural networks to suggest more robust models [19, 20]. In [21], a method is presented for creating attractors during the training of an iterated autoencoder.

Autoencoders have been used for OOD detection before. The basic principle behind this approach is that an autoencoder can accurately reconstruct ID samples but fails to reconstruct OOD samples during testing [24]. However, recent studies have shown that in some cases, autoencoders can also reconstruct OOD samples more effectively than ID samples [25]. Studies have been done to develop robust OOD detection methods, but they often lose their performance under high perturbation levels [5,10]. These studies merely concentrate on enhancing the ability to detect OOD inputs in adversarial examples, and the proposed methods do not help make the main classification more robust. Even in some cases, there is a tradeoff between improving the robustness of OOD detection and maintaining the accuracy of the main classification [23].

Considering this gap, we propose in this study to train samples as attractors in an iterated deep autoencoder network and use the reconstruction results to detect OOD. The autoencoder learns a basic representation of the ID dataset and reconstructs it with minimum error. Therefore, when the reconstruction error is high and the similarity between the input and its reconstruction by the autoencoder is low, the input is susceptible to being an OOD example. In this method, the feedback connection and iteration over the network will help us solve the problem of falsely predicting OOD as an ID example in deep autoencoder networks. By training ID samples as attractors of the deep autoencoder network, the network will handle a wide range of input variations and perturbations. In our work, we investigate both adversarial ID and adversarial OOD examples. Simultaneously with the OOD detection method, we improve the main classifier's robustness and the overall accuracy of the classification task among adversarial examples. We evaluate the proposed method's robustness and accuracy by testing it against adversarial examples generated by FGSM attacks, specifically focusing on high perturbation levels. The main contributions of this research can be briefly outlined as follows:

- Introducing the attractor network for classification and OOD detection. We propose an architecture for this network and explain the training process that contains a loss function suitable for training samples as the network's attractors.

- Selecting a function to calculate the similarity score between input and its reconstruction using the attractor network. It is shown that this function is capable of distinguishing between ID and OOD samples accurately.

- Robustness analysis of the attractor network in terms of how this network maintains its performance in classification and OOD detection under adversarial attacks. We demonstrate that the attractor network, by using the concept of attractors in neural networks, can handle high levels of perturbations.

# 2 Materials and methods

## 2.1 Using Deep Autoencoder Network for OOD Detection

Reconstruction-based methods rely on the concept that the encoder-decoder model, which is trained on ID data, produces distinct results for ID and OOD samples. This difference can be captured in either the latent feature space or the pixel space of reconstructed images. Reconstruction models like deep autoencoder networks, trained only on ID data, may not be able to recover OOD data properly. This happens because the OOD sample may contain some unrecognized features for the trained autoencoder, leading to the filtering of unknown components during encoding. When the autoencoder reconstructs the input, it does so without including all the components of the latent feature space. This results in a reconstructed output that differs from the original input and the OOD sample can be identified. Thresholding on the differences between input and its reconstruction, such as reconstruction error, can be used for classifying between ID and OOD samples [25].

## 2.2 Robust OOD Detection

Although current techniques for detecting OOD samples have yielded promising results on natural ID and OOD samples, several studies have shown their vulnerability to variations, particularly adversarial attacks [9,10,11]. This means that adversarial inputs created by such attacks can deceive the OOD detectors and reduce their performance. Therefore, it is crucial to develop robust OOD detection methods that maintain their performance under any perturbations, especially these attacks.

### 2.2.1 Adversarial Attacks

In the testing stage, perturbations applied to a trained model's inputs can yield distinct outputs from the original inputs. A small perturbation can cause the input to move out of its true class's decision region and into another decision region in the subspace learned by the neural network. This means that the perturbations can cause the input to cross the decision boundaries previously learned by the model, as shown in Figure 1.

This section delves into an adversarial attack algorithm that reveals the susceptibilities of neural network classifiers and OOD detection methods. This research uses a gradient-based adversarial attack called Fast Gradient Sign Attack (FGSM), which is highly effective in misleading neural networks. FGSM is a widely used adversarial attack introduced by Ian Goodfellow et al. [14]. This white-box method targets neural networks after they have been trained by exploiting their training process, specifically gradients. The attack uses the gradient of the input data's cost function and then adjusts the input data to maximize the cost function by intentionally adding noise. Equation (1) illustrates how this attack produces an adversarial example.

$$X_{adv} = X + \epsilon * sign(\nabla_x J(X, y_{true})) \quad (1)$$

In this equation, $X$ is the clean image, $X_{adv}$ is the adversarial image, $\epsilon$ is the size of the adversarial perturbation, which is always specified in terms of pixel values within the range of 0 to 255, $\nabla_x$ calculates the gradient with respect to x, and the cost function used to train the model is denoted as $J(X, y_{true})$.

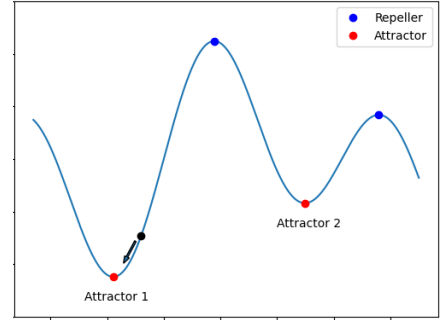

**Figure 2:** Dynamic system with attractors and repellers. Red points, shown as valleys, indicate stable points or attractors. Blue points, shown as peaks, indicate unstable points or repellers. A ball (black point) rolling on this landscape will find the most stable positions at the bottoms of the valleys.

## 2.3 Attractors Used to Improve Robustness

According to the theory of self-organization, a system of interconnected elements such as neural networks creates order around attractors that help to establish and maintain stable patterns within the system. These attractors create a landscape that shapes and defines the system's dynamic [30]. Figure 2 shows a simple landscape of a dynamic system, with colored points indicating its fixed points. The red points represent the stable points or attractors of the system, which are depicted as valleys. On the other hand, the blue points represent the unstable points or repellers, which are depicted as peaks. If we picture a ball (the black point) rolling across this landscape, we can easily see that the bottoms of the valleys present the most stable positions for the ball.

Attractors are incredibly valuable in artificial neural networks, particularly in pattern recognition. We can create an iterated system with dynamic behavior by introducing a feedback connection from the network's output to its input. This system can be expressed as a first-order recurrent relationship, as shown in Equation (2). In this equation, x represents input, and f is assumed to be a function of a deep neural network. After several iterations, when the output value remains constant, the fixed points of the neural network can be determined. When k → ∞ and the system's output remain constant, $x_{k+1}$ equals $x_k$, indicating that $x_k$ is a fixed point. This method is widely used to determine a system's fixed points.

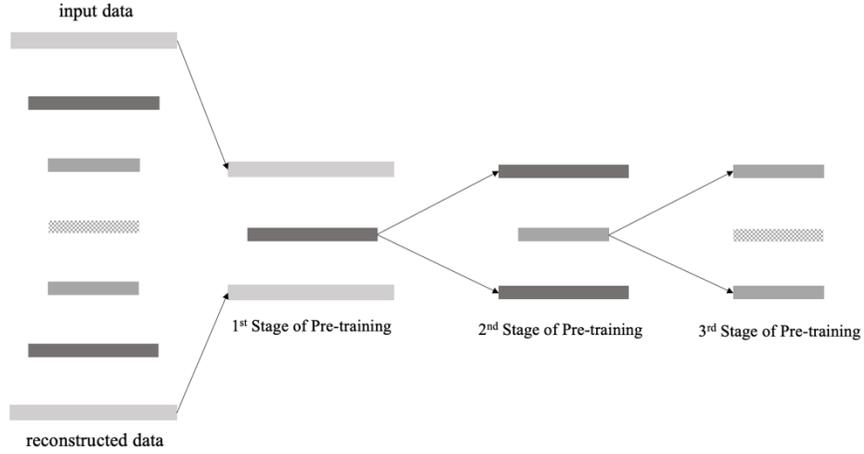

**Figure 3:** Stages of layer-by-layer pre-training. Each stage demonstrates breaking down the deep autoencoder network into several single hidden-layer autoencoders.

$$x_{k+1} = f(x_k) \quad (2)$$

As previously mentioned, a fixed point can either be stable or unstable. A study in [29] provides a stability analysis of fixed points to differentiate attractors. According to this study, If we assume that $x^*$ represents the fixed point of the system and f is continuously differentiable in the neighborhood of $x^*$, then $x^*$ is referred to as a stable fixed point or attractor if $|real(f'(x*))| < 1$; Otherwise, it is referred to as an unstable fixed point. Equation (3) summarizes these conditions to determine the attractors.

$$x^* \text{ is a } \begin{cases} stable\ point,\ if\ |real(f'(x^*))| < 1 \\ unstable\ point,\ otherwise \end{cases} \quad (3)$$

In a pattern recognition problem, all desired patterns can be trained as attractors of the deep neural network. When the network is given an input pattern as the starting point, it goes through several iterations and eventually converges to the corresponding attractor pattern [31]. Interestingly, even if some variations appear in the input, the system will still converge to the corresponding attractor due to its attractor properties. In simple terms, attractors can handle perturbed inputs and recognize them more successfully, thus making the recognition process more robust [32]. According to this, attractors can be utilized to develop robust classification methods.

## 2.4 Pretraining Deep Autoencoder Networks

Pre-training is a technique that can help deep neural networks avoid getting stuck in local minima during training, which ultimately allows them to converge to a solution faster. A study conducted by He et al. [27] found that pre-training and training from scratch can lead to similar accuracy. However, it's worth noting that this only holds true for unperturbed data, as demonstrated by Hendrycks et al. [26]. The benefits of pre-training go beyond quick convergence; it can also improve model robustness and uncertainty, leading to better handling of variations and perturbations.

In this research, we use the layer-by-layer pre-training method [28]. This method adjusts the hyperplanes among the training data by modifying the weights to maximize discrimination between them. Figure 3 illustrates this method's performance. This approach involves breaking down a complex and deep network into a group of one-hidden-layer autoencoders. During the pre-training stage, each network is trained separately. Later, in the next stage, all of these networks are combined for final fine-tuning.

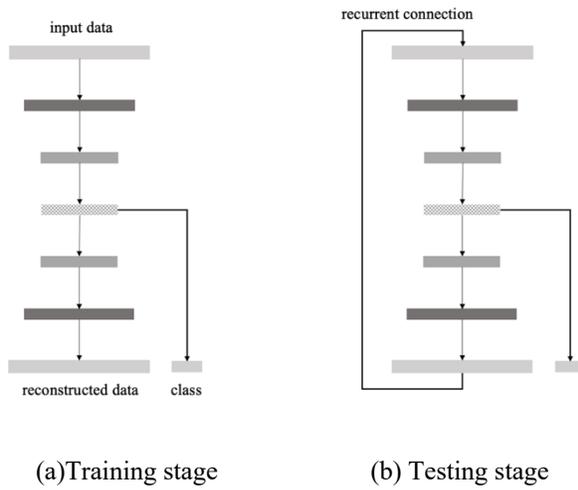

**Figure 4:** The proposed structure of the deep attractor neural network. (a) The structure utilized during the training stage, (b) The structure employed during the testing stage, which includes a feedback loop in addition to the structure used in the training stage.

## 2.5 Proposed Method

### 2.5.1 Structure of the Deep Attractor Neural Network

In the preceding sections, we presented some introductory materials. In this section, we will elaborate on the proposed method of this research. Firstly, we will introduce the structure of the neural network we have employed for robustly detecting OOD samples and classifying inputs.

The autoencoder used in this research consists of fully connected layers with 784, 500, 300, 300, 300, 500, and 784 neurons, respectively, assuming a 28x28 input. Additionally, a connection from the autoencoder's bottleneck layer to the 10-neuron layer is made for the 10-class classification task, as demonstrated in Figure 4. The autoencoder layers use the ReLU activation function, and the classification layer uses the Softmax activation function.

In this structure, the network reconstructs the input pattern and classifies it simultaneously rather than separately. The autoencoding part of the structure aims to minimize the reconstruction error. In contrast, the classification part tries to classify the input with the most minor error based on the features extracted in the bottleneck layer. By designing the structure in this way, the network learns to extract features from the input in the bottleneck layer that are both discriminative between different classes and contain important input information for reconstruction.

In the training stage, the network learns identity mapping in the structure shown in Figure 4-a. According to Equation (3), this mapping ensures that the input pattern is learned as a fixed point. Ideally, all input patterns in the training stage should be learned as stable fixed points or attractors. During the testing stage, a feedback loop is established that connects the output of the autoencoder to its input, creating a recurrent relation (Figure 4-b). This means the system's output is fed back to the input, and this loop continues infinitely. The system has reached the corresponding fixed point when the output of the system remains unchanged, indicating that it will not change in the subsequent iterations. This procedure helps the input pattern converge to the corresponding template pattern learned in the training stage as an attractor, which is helpful in canceling the possible perturbations and preventing the network from getting deceived by them.

### 2.5.2 Training the Proposed Network

In this study, the network is divided into small one-hidden-layer autoencoders using the layer-by-layer pre-training approach. During pre-training, three loss functions, namely $L_{reconstruction}$, $L_{eigen}$, and $L_{classification}$, are employed in every stage. As shown in Equation (4), the total loss function will be constructed of the weighted sum of the three loss functions mentioned.

$$L_{total} = \alpha_1 L_{reconstruction} + \alpha_2 L_{eigen} + \alpha_3 L_{classification} \quad (4)$$

In this Equation, $\alpha_1$, $\alpha_2$, and $\alpha_3$ determine the influence of each loss function during training. $L_{reconstruction}$ is the mean squared error loss function aimed at minimizing the reconstruction error of the autoencoder. L_classification, on the other hand, is the cross-entropy loss function intended to enhance the accuracy of input classification in the proposed autoencoder network. L_eigen is a loss function based on eigenvalues, designed to maximize the number of attractors from training data [21]. It imposes a constraint on the eigenvalues of the Jacobian matrix to keep them as small as possible. According to Theorem 1, this loss function causes training data to be attractors of the network rather than repellers. The cost function calculation based on the eigenvalues is observed in Equation (5), where $a_{ij}$ represents the elements of the Jacobian matrix. As observed, this cost function calculates the output relative to the input by halving the sum of squares of the elements of the Jacobian matrix.

$$L_{Eigen} = \frac{1}{2} \sum_{i=1}^{n} \sum_{j=1}^{n} a_{ij}^2 \quad (5)$$

Following pre-training, all components are integrated for final fine-tuning. It's worth noting that all three loss functions also persist in the fine-tuning. We refer to this training approach as the attractor training.

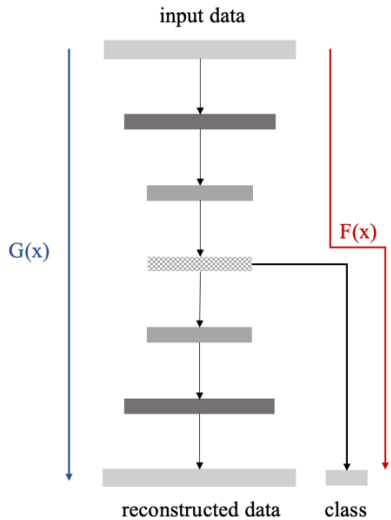

**Figure 5:** F(x) and G(x) in the proposed deep attractor neural network. F(x) represents the classification function, and G(x) refers to the autoencoding function.

## 2.5.3 OOD Detection Using Deep Attractor Neural Network

Studies have shown that an autoencoder network can reconstruct data it has learned. This leads to a better reconstruction of ID examples than OOD examples, so we can use the reconstruction performance to detect OOD samples. In some cases, the trained network can also effectively reconstruct OOD samples, making it difficult to differentiate between ID and OOD samples using the reconstruction performance. This research utilizes a deep autoencoder network that incorporates a classification component. This enables the extracted features from the input, compressed in the bottleneck layer, to be important for reconstruction and distinguishable between different classes.

By emphasizing these distinguishing features existing in ID samples, the network's reconstruction of OOD samples is expected to be weaker, resulting in better OOD detection. Moreover, this study considers training ID samples as attractors so that if adversarial attacks or other variations perturb the input, it can be attracted to the corresponding attractor in its basin. As shown in Figure 4-a, the network not only identifies OOD samples but also classifies inputs recognized as ID samples, making this method more comprehensive. By learning ID samples as attractors, the classification and OOD detection processes are expected to become more robust against perturbations and variations.

The approach utilized in this research to evaluate reconstruction performance and identify OOD examples is based on computing the similarity between the input and output. The autoencoder produces the output by reconstructing the input, which may require additional iterations to remove perturbations and attain attractors. If the similarity is greater than a certain threshold, the input is classified as an ID example. On the other hand, if the similarity is smaller than the threshold, the input is classified as an OOD example. The algorithm for this method is shown in Equation (6), where x represents the input, G refers to the autoencoder function (as shown in Figure 5), D denotes the OOD detection function, T is the similarity threshold, and n is the number of iterations needed to reach the attractor.

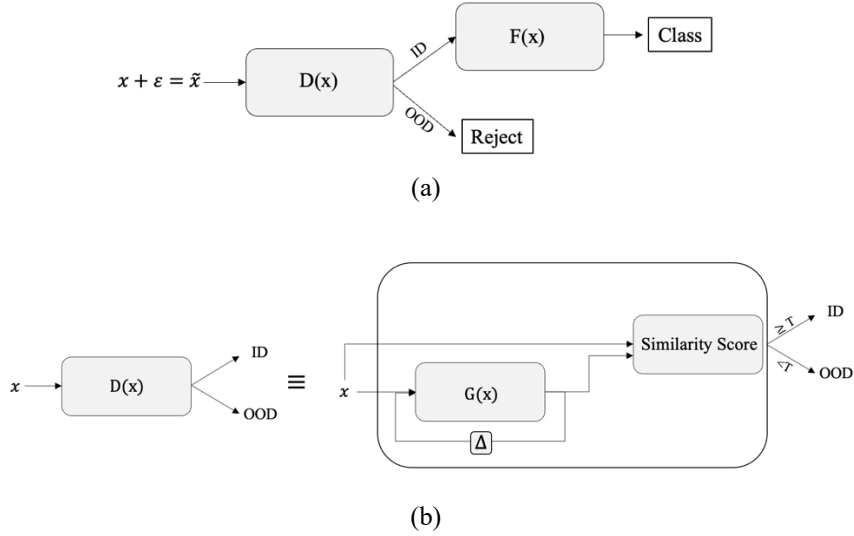

**Figure 6:** Diagram of classification in the presence of OOD and adversarial samples. (a) Rejecting OOD inputs and classifying only ID inputs, (b) Proposed OOD detection method.

$$D(x) = \begin{cases} 1, & if \quad \text{Similarity}(\underbrace{G \circ G \circ \ldots G(x)}_{n}, x) \geq T \\ 0, & if \quad \text{Similarity}(\underbrace{G \circ G \circ \ldots G(x)}_{n}, x) < T \end{cases} \quad (6)$$

Figure 6 demonstrates the process of classifying input in the presence of OOD examples and adversarial examples. Upon input is received, it undergoes D(x) to identify OOD examples. OOD examples are rejected, and only ID inputs are classified by F(x), as illustrated in Figure 6-a. Figure 6-b illustrates the specifics of D(x), as explained earlier. To determine the similarity score, the input is compared to the output generated by G(x). In certain instances, it may be necessary to iterate through G(x) multiple times to obtain a clean pattern, as indicated by the feedback loop around G(x).

### 2.5.4 Calculation of Similarity Score

To calculate the similarity between the input and output of the proposed network, there are many methods available, such as calculating the sum of squares of differences, Euclidean distance, Mahalanobis distance, and so on; the Pearson correlation coefficient serves as a valuable measure for determining the similarity quantity between two images when they are transformed into flat vectors. Pearson correlation evaluates the linear relationship between corresponding elements in these flat vectors by representing each image as a pixel intensity vector. This study uses the Pearson correlation coefficient to calculate the similarity between the input and reconstructed image. The advantage of using this similarity function over using reconstruction error is that it has less strictness in measuring similarity, and due to normalization using image variances, the results are comparable. The Pearson correlation coefficient is calculated according to Equation (7).

$$r(x,y) = \frac{\sum_{i=1}^{n}(x_i - \bar{x})(y_i - \bar{y})}{\sqrt{\sum_{i=1}^{n}(x_i - \bar{x})^2 \sum_{i=1}^{n}(y_i - \bar{y})^2}} \quad (7)$$

In this equation, $x_i$ and $y_i$ represent the pixels of two images, and $\bar{x}$ and $\bar{y}$ denote their means. The Pearson correlation coefficient close to 1 shows a strong positive linear relationship, indicating high similarity between the images.

# 3 Results & Discussion

## 3.1 Experiments

In this section, we perform extensive experiments to evaluate our classification method under adversarial attacks. Our main findings are summarized as follows:

- The proposed reconstruction-based OOD detection method, without the presence of $L_{eigen}$ in the training process, can accurately detect OOD samples and classify ID samples among clean images. However, its performance degrades under adversarial attacks.

- By learning training samples as its attractors, the network can preserve crucial information about the input image under perturbations, making the adversarial attack against it unsuccessful.

- Training the samples as the network's attractors enhances the robustness of the network significantly against adversarial attacks.

We provide further information regarding the experiments in the following.

### 3.1.1 Datasets
In this study, three reputable datasets in the field of deep learning, namely MNIST [33], Fashion-MNIST [34], and CIFAR-10 [35], have been utilized. MNIST contains grayscale images of handwritten digits and is composed of a training set with 60,000 samples and a test set with 10,000 samples. Each image is classified into one of 10 categories and has a size of 28 by 28 pixels.
Fashion-MNIST is a dataset of fashion-related images with the same sample size and classification as MNIST. Similarly, CIFAR-10 is a dataset with 60,000 RGB images categorized into ten different classes. To suit the network used in this study, the CIFAR-10 images were resized to 28 by 28 pixels, and only the first channel of the image was utilized. This dataset is referred to as CIFAR-10-bw in this research.

In experiments conducted, the MNIST dataset is used as the ID sample and training data. Also, the Fashion-MNIST and CIFAR-10-bw datasets are used as OOD samples, and 10000 samples of each dataset are involved in the testing stage.

### 3.1.2 Training Configurations
We previously explained in detail the architecture and training process of the neural network we employed in this study. To evaluate the impact of training ID samples as attractors of the network, we considered two training methods: baseline training and attractor training. The resulting networks are referred to as the baseline network and the attractor network, respectively. The attractor training was explained before (section 2.5.2). This training uses all three loss functions ($L_{reconstruction}$, $L_{eigen}$, and $L_{classification}$). On the other hand, baseline training just involves two of these loss functions in training, $L_{reconstruction}$ and $L_{classification}$. The $L_{eigen}$ is also employed in the attractor training, which compels training samples to be the network's attractors.
In all stages of pretraining and fine-tuning, the Adam optimizer has been employed, with a batch size of 64. During the three pre-training stages and fine-tuning, the learning rates used were 0.001, 0.005, 0.005, and 0.0005, respectively. The number of training epochs for all stages was set to 100, along with early stopping (based on a criterion of accuracy reduction over eight consecutive epochs on the validation data).

### 3.1.3 Hyperparameters
According to Equation (4), the contribution of each loss function to the total loss function is determined by the coefficients $\alpha_1$, $\alpha_2$, and $\alpha_3$. In attractor training, these coefficients are set to 0.988, 0.002, and 0.01, respectively. However, in baseline training, $L_{Eigen}$ is not used, so the value of $\alpha_2$ is zero, and the values of $\alpha_1$ and $\alpha_3$ are set to 0.99 and 0.01, respectively.
The FGSM method generates adversarial inputs based on a single parameter, namely ε, which denotes the magnitude of the perturbation, and in this study, it ranges from 1/255 to 10/255.

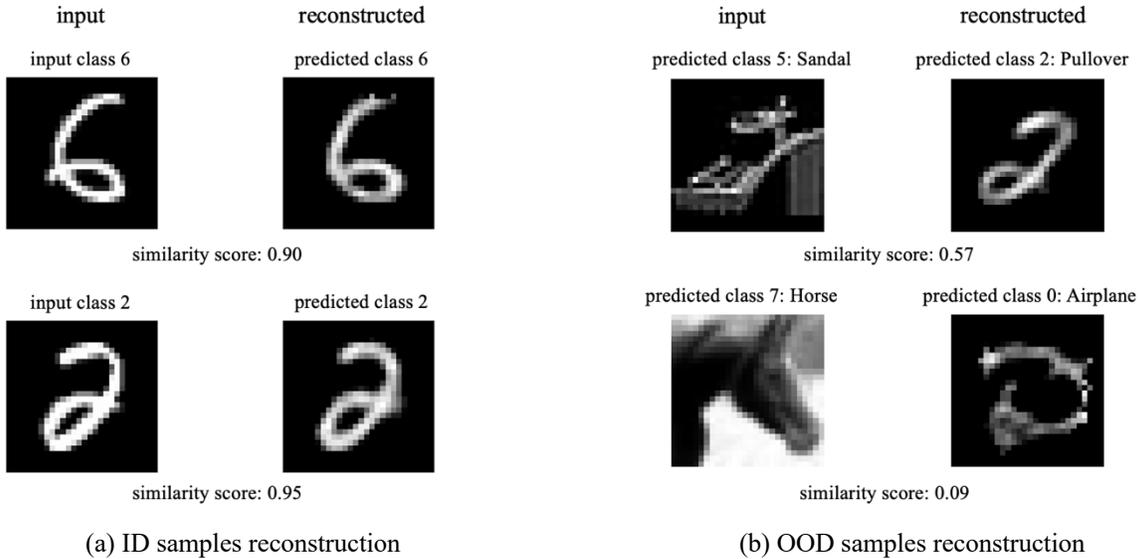

**Figure 7:** Reconstruction of inputs by the baseline network. Each row of each column is dedicated to an input reconstruction, and the similarity score between the input and its reconstruction is reported below. (a) Inputs are ID examples from the same distribution of training data, (b) Inputs are OOD examples from a different distribution than the training data.

## 3.2 Test Results

### 3.2.1 Baseline Network

For this part, we trained the proposed network using the baseline training method and the MNIST dataset, which is referred to as the baseline network. The accuracy of the classification of MNIST test data, which is unseen during training, is 98.17%. In the next step, we tested the reconstruction hypothesis, which claims that the trained autoencoder network can reconstruct ID samples with minimum error.

Figure 7-a displays inputs selected from MNIST test data on the left side, while their reconstruction by the baseline network is shown on the right. The network has effectively reconstructed the inputs from the same distribution of training data with minimal error while accurately predicting their class as indicated on top of each image. A similarity score is computed between the inputs and their corresponding reconstructed images, which is presented below each example in Figure 7-a and shows high levels of similarity. On the other hand, inputs in Figure 7-b are from different distributions selected from fashion-MNIST and CIFAR-10-bw datasets. Low accuracy in the reconstruction and classification of OOD examples is seen in the baseline network trained over MNIST data, leading to low similarity between input and reconstructed images.

We conducted a comparison to determine whether the proposed network can differentiate between ID and OOD examples. We calculated similarity scores between the input and its reconstructed image, as shown in Figure 8-a. Firstly, we calculated the similarity scores for 10000 ID samples from the MNIST test dataset. These samples were not seen during the training phase and are represented by blue points. Additionally, we plotted the similarity scores between 10000 OOD samples from the fashion-MNIST test dataset in Figure 8-a as red points. It is apparent that the distribution of calculated similarity scores differs between ID and OOD samples, and we can distinguish ID and OOD examples by demonstrating thresholding (the black line) on similarity scores.

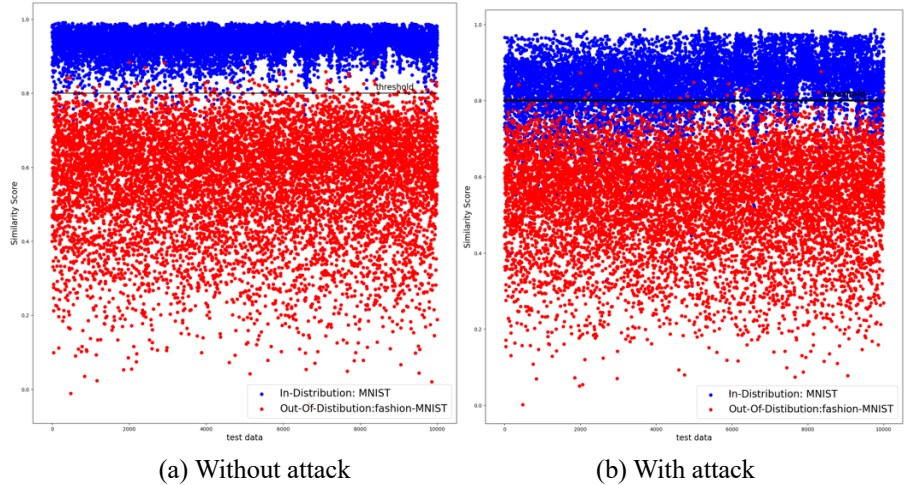

(a) Without attack  (b) With attack

**Figure 8:** The similarity scores between input and its reconstruction using the baseline network for ID (blue points) and OOD samples (red points). ID and OOD samples are selected from MNIST and Fashion-MNIST test datasets, respectively. (a) In the absence of adversarial attacks, (b) In the presence of FGSM attack with ε=20/255.

**Table 1:** The evaluation of classification between ID and OOD samples using the baseline network.

| Datasets | Accuracy | Precision | Recall | Accuracy | Precision | Recall |
|---|---|---|---|---|---|---|
| | Without attack | | | With attack (FGSM, ε=20/255) | | |
| $D_{in}$ = MNIST $D_{out}$ = fashion-MNIST | 99.15% | 99.15% | 99.15% | 88.84% | 99.67% | 77.93% |
| $D_{in}$ = MNIST $D_{out}$ = CIFAR-10-bw | 99.50% | 99.84% | 99.15% | 88.93% | 99.91% | 77.93% |

Based on the observations and Equation (6), a binary classification was performed to distinguish between ID (class 1) and OOD (class 0) samples. This was accomplished by setting T (threshold) to 0.8, which allowed for the detection of OOD samples. This approach helped to avoid misclassification by rejecting the OOD samples in the classification process, as the trained network lacked the knowledge to make decisions about these samples. The results of this OOD detection method are reported in Table 1. 10000 ID samples were utilized, which are from the MNIST test dataset, and 10000 OOD samples were employed from each fashion-MNIST and CIFAR-10-bw dataset. The classification is evaluated by calculating accuracy, precision, and recall. The higher the values of these metrics, the better the classification performance. Since the classification results showed high measures, the OOD samples were detected accurately.

To study the influence of adversarial attacks on OOD detection and classification using the baseline network, we applied adversarial perturbations on both ID and OOD samples using FGSM attack with ε=20/255. The accuracy of the baseline network in classifying perturbed ID samples was **27.84%**, significantly lower than the 98.17% accuracy achieved with clean samples.

Figure 9 compares the effect of random noise and FGSM adversarial attack on the classification of an ID sample. Figure 9-a shows an ID sample without any perturbations belonging to class 5. As mentioned on top of the image, the baseline network has accurately predicted the actual class of this clean input. In Figure 9-b, we can see that the baseline network is highly resistant to salt and pepper random noise applied with 0.2 probability to the demonstrated clean image, resulting in accurate classification. Figure 9-c shows the baseline network failing to classify the ID

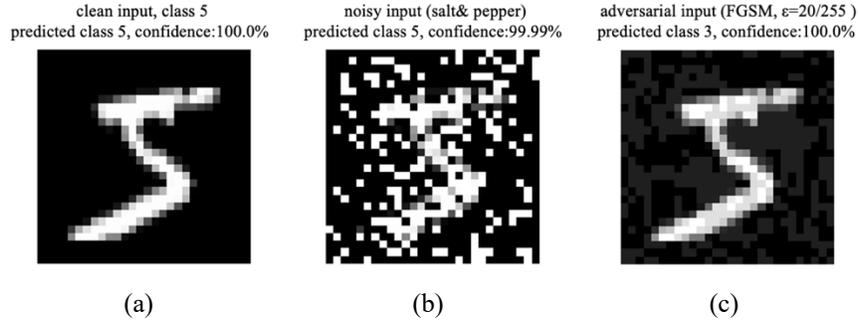

**Figure 9:** Effectiveness of adversarial attack in reducing classification performance. Classification of (a) input without perturbation, (b) input perturbed with random noise, and (c) input perturbed with adversarial attack.

adversarial example generated by the FGSM attack with ε=20/255 on the clean image.

It is observed from Figure 9 that although the level of perturbations in the image contaminated with salt and pepper noise is much higher than the level of perturbations in the image generated by the adversarial attack, the neural network can still correctly classify the class of the former due to the random nature of noises applied to the image. However, it fails to classify the adversarial input correctly. This highlights the high effectiveness of adversarial attack methods in fooling neural networks compared to other variations.

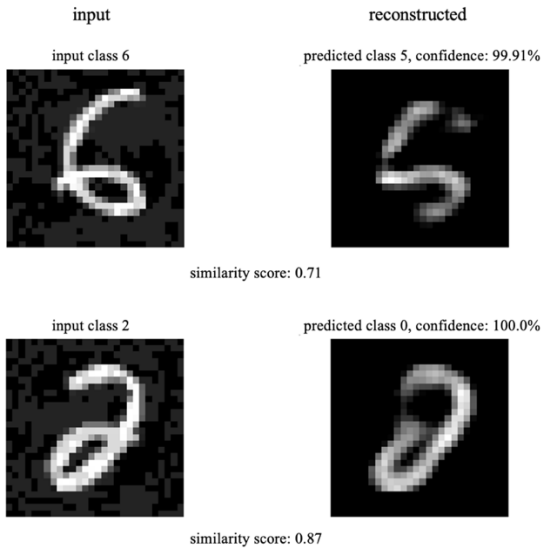

**Figure 10:** Reconstruction of ID adversarial inputs by the baseline network. In each row, the predicted class of input and similarity score between input and reconstructed image are reported.

In addition to the classification accuracy of ID adversarial examples, their reconstruction performance using the baseline network has also been investigated. Figure 10 displays the poor performance of the reconstruction process on samples that were generated as adversarial examples of clean images shown in Figure 7-a. The similarity scores between input and reconstructed images have decreased in all demonstrated cases due to weaker reconstruction, as mentioned below each example. Analogous to Figure 8-a, we compared similarity scores between ID and OOD samples, this time in the presence of adversarial attacks. As shown in Figure 8-b, in the presence of adversarial attacks, ID and OOD samples are hard to distinguish, and previous thresholding depicted as the black line is no longer appropriate for accurate classification between ID and OOD adversarial samples. Our assessment of the classification of ID and OOD adversarial examples generated by FGSM attack with ε=20/255, using the baseline network, is outlined in Table 1 and shows lower performance than the previous experiment in the absence of any adversarial attacks.

### 3.2.2 Attractor Network

In this section, we utilized the attractor training method to train the proposed network. We employed 1000 samples from the MNIST dataset to train the attractor network. This particular training method has the advantage of convergence in fewer epochs, but the downside is that it comes with a high computational cost. This is because the Jacobian matrix needs to be computed during each training epoch. As a result, we needed to find a balance between the computational costs and the improved network's performance. However, we found 1000 training samples to be sufficient for our purposes.

In the previous section, we used 50000 training samples for baseline training. However, in the current section, we have reduced the number of training samples by 50 times. Despite the reduction in training data, the attractor network was still able to accurately classify the MNIST test data with an accuracy of 88.73%. Even when the same test data was perturbed, the classification accuracy only dropped slightly to **87.13%**, which is still quite close to the accuracy obtained when the samples were clean. This demonstrates that the attractor network is robust against adversarial attacks and maintains its performance under perturbations. In contrast, when a similar experiment was conducted using the baseline network, classification accuracy dropped significantly to 27.84%. Therefore, the attractor network is more suitable than the baseline network for classifying adversarial samples.

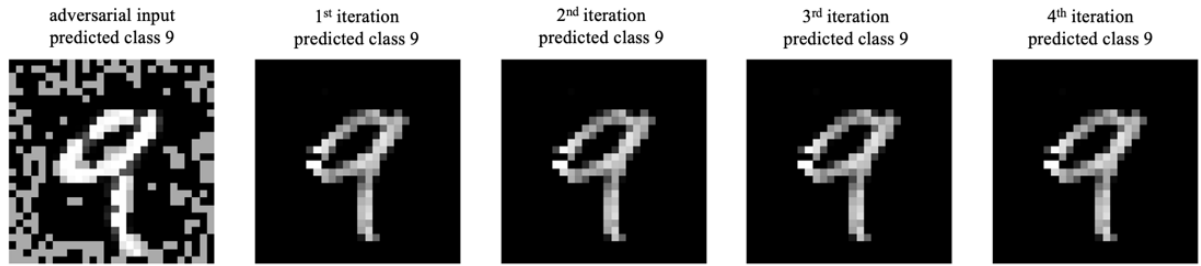

**Figure 11:** the attraction of an adversarial training sample to its corresponding clean image. The image on the left is the adversarial input generated using the FGSM method with ε=255/255. Furthermore, the outputs of successive iterations in the attractor network are presented. (after 1st iteration, the output does not change considerably)

In Figure 11, to assess whether a training sample has been learned as an attractor of the trained network, it has been perturbed using the FGSM attack with ε=255/255 and fed into the network as its input. According to the architecture of the attractor network shown in Figure 4-b, the input has been iterated through the network, and the reconstructed image at each iteration is represented. As can be seen, the perturbed sample is attracted to its corresponding clean image in the first iteration and remains constant after the subsequent iterations. This indicates that this sample is successfully learned as an attractor of the network during training, and this network can help adversarial samples with a high level of perturbations converge to their corresponding clean images.

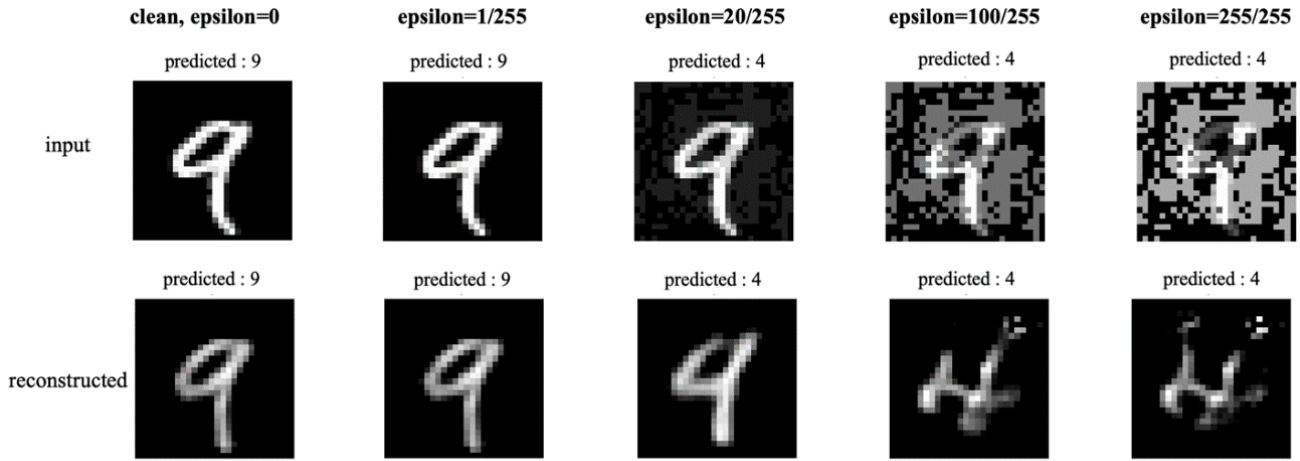

(a) Results of the baseline network

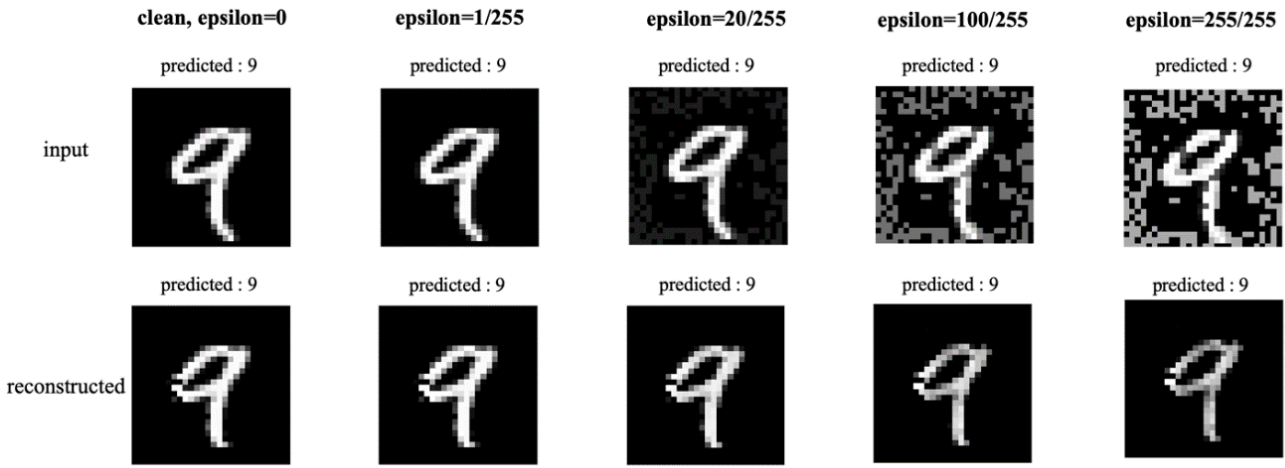

(b) Results of the attractor network

**Figure 12:** The reconstruction and classification performance in different levels of perturbation using (a) the baseline network and (b) the attractor network.

Figure 12 shows the performance difference between the baseline network and the attractor network in reconstructing and classifying a training sample. As depicted in Figure 12-a, the baseline network accurately reconstructed and classified the sample in the absence of adversarial attacks. However, as the image perturbation level increased by raising the value of ε in the FGSM attack, both the reconstruction and classification errors increased significantly. Conversely, according to Figure 12-b, the attractor network maintained good performance in both the reconstruction and classification of the input, even under high perturbation levels.

The important point is the difference between adversarial attacks on the baseline network and the attractor network, both with the same level of perturbation. The FGSM attack uses the cost function's gradient of the trained model to add perturbation to the sample and deceive the learning model. Since the cost function used for training the baseline network and the attractor network are different, adversarial inputs generated using the cost function's gradient of these networks are different, as shown in the first row of Figure 12-a and Figure 12-b. This example demonstrates that the attractor network preserves crucial information about the input image, making the adversarial attack unsuccessful against it.

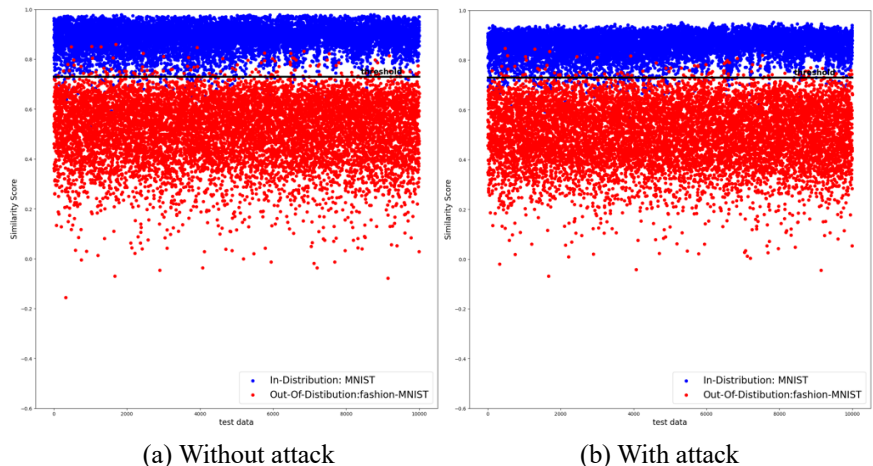

(a) Without attack  (b) With attack

**Figure 13:** The similarity scores between input and its reconstruction using the attractor network for ID (blue points) and OOD samples (red points). ID and OOD samples are selected from MNIST and Fashion-MNIST test datasets, respectively. (a) In the absence of adversarial attacks, (b) In the presence of FGSM attack with ε=20/255.

**Table 2:** The evaluation of classification between ID and OOD samples using the attractor network.

| Datasets | Accuracy | Precision | Recall | Accuracy | Precision | Recall |
|---|---|---|---|---|---|---|
| | Without attack | | | With attack (FGSM, ε=20/255) | | |
| $D_{in}$ = MNIST $D_{out}$ = fashion-MNIST | 98.84% | 98.67% | 99.01% | 98.48% | 98.90% | 98.04% |
| $D_{in}$ = MNIST $D_{out}$ = CIFAR-10-bw | 99.28% | 99.55% | 99.01% | 98.88% | 99.71% | 98.04% |

Figure 13 depicts the similarity scores between input and its reconstruction using the attractor network for ID (blue points) and OOD samples (red points). Similar to Figure 8, ID and OOD samples are selected from MNIST and Fashion-MNIST test datasets, respectively. In Figure 8-a, these samples are used in their original state, but in Figure 8-b, samples are perturbed by FGSM attack with ε=20/255. This example illustrates that the attractor network can accurately differentiate between ID and OOD samples by using the chosen similarity score and threshold despite high levels of perturbation generated by adversarial attacks. Meanwhile, Table 2 outlines the assessment of the attractor network's classification of ID and OOD samples, both in the absence and presence of the FGSM adversarial attack. The results show the good performance of the attractor network in OOD detection, even when facing adversarial samples.

The experiments we conducted in this study showed good performance of the baseline and attractor networks in the classification and OOD detection of normal samples. However, in the presence of adversarial attacks, the baseline network lost its performance in both classification and OOD detection. Nonetheless, the attractor network preserved its performance even in high levels of perturbation. Since the only difference between these two networks is the presence of $L_{Eigen}$ in the total loss function utilized for training the networks, we can infer that this term in loss function improves the robustness against adversarial attacks in both classification and OOD detection tasks.

# 4 Conclusion

In conclusion, this study presents a unique and innovative method for detecting out-of-distribution (OOD) samples using neural networks. By leveraging the concept of attractors within neural networks, the proposed approach demonstrates robustness against adversarial attacks. Crucially, it not only identifies and removes OOD samples but also effectively categorizes the remaining inputs as In-distribution (ID) samples using a single network, even under significant perturbations. The method's resilience to adversarial attacks underscores its potential as a promising defense strategy for future applications.

The prevalence of OOD and adversarial examples in input data presents significant challenges for deploying neural networks in real-world scenarios. This study addresses these challenges by enhancing the reliability of neural networks. By providing a means to filter out irrelevant inputs lacking sufficient information for decision-making, the method contributes to creating more dependable neural networks. Moreover, its ability to restore a clean form of input from perturbed ones further enhances network effectiveness and reliability, making it a valuable tool for practical applications.

Additionally, it is crucial to focus on refining the loss function to mitigate computational costs associated with implementing the method at scale. Looking ahead, future research should explore this OOD detection method across various datasets and tasks to better understand its performance and reliability. Such investigations will facilitate its integration into real-world applications with confidence.

**CRediT authorship contribution statement**
Nasrin Alipour: Conceptualization, Investigation, Methodology, Software, Visualization, Writing – review & editing. Seyyed Ali Seyyedsalehi: Conceptualization, Supervision, Writing – review & editing.

**Declaration of Competing Interest**
On behalf of the corresponding author, all authors declare that they have no conflict of interest.